\documentclass[acmsmall,screen]{MapleTrans}
\usepackage{url}            
\usepackage{booktabs}       
\usepackage{amsfonts}       
\usepackage{nicefrac}       
\usepackage{microtype}      
\usepackage{lipsum}
\usepackage{fancyhdr}       
\usepackage{graphicx}       
\graphicspath{{media/}}     
\usepackage{ragged2e}
\usepackage{amsmath}
\usepackage{mathrsfs}
\usepackage[nocompress]{cite}
\usepackage[numbers,sort&compress]{natbib}
\newcommand{\upcite}[1]{\textsuperscript{\textsuperscript{\cite{#1}}}}
\usepackage[normalem]{ulem}

\usepackage{enumitem}
\usepackage{etoolbox}
\usepackage{indentfirst}
\usepackage{breqn}
\usepackage{lineno}
\usepackage{maple}
\usepackage{xfrac}
\usepackage{xcolor}
\definecolor{mygreen}{rgb}{0,0.6,0}
\definecolor{mygray}{rgb}{0.5,0.5,0.5}
\definecolor{mymauve}{rgb}{0.58,0,0.82}
\definecolor{altblue}{rgb}{0.0,0.6,1.0}
\definecolor{lstbg}{cmyk}{0.05, 0.01, 0, 0}
\definecolor{morebluish}{cmyk}{0.06,0.04,0,0}
\usepackage{listings}
\input{listings-maple-definition.sty}
\begin{document}

\title[~]{Random Graph-Based Neuromorphic Learning with a Layer-Weaken Structure\\
(This work has been submitted to the IEEE for possible publication. Copyright may be transferred without notice, after which this version may no longer be accessible)}

\author{Ruiqi Mao}
\email{ruiqimaonwpu@163.com}
\orcid{~}
\author{\textbf{Rongxin Cui}}
\authornotemark[1]
\email{r.cui@nwpu.edu.cn}
\affiliation{%
  \institution{School of Marine Science and Technology, Northwestern Polytechnical University}
  \streetaddress{West Youyi Road}
  \city{Xi'an}
  \state{Shaanxi}
  \country{P. R. China}
  \postcode{710072}
}

\renewcommand{\shortauthors}{Ruiqi Mao and Rongxin Cui}

\begin{abstract}
\textbf{Unified understanding of neuro networks (NNs) gets the users into great trouble because they have been puzzled by what kind of rules should be obeyed to optimize the internal structure of NNs. Considering the potential capability of random graphs to alter how computation is performed, we demonstrate that they can serve as architecture generators to optimize the internal structure of NNs. To transform the random graph theory into an NN model with practical meaning and based on clarifying the input-output relationship of each neuron, we complete data feature mapping by calculating Fourier Random Features (FRFs). Under the usage of this low-operation cost approach, neurons are assigned to several groups of which connection relationships can be regarded as uniform representations of random graphs they belong to, and random arrangement fuses those neurons to establish the pattern matrix, markedly reducing manual participation and computational cost without the fixed and deep architecture. Leveraging this single neuromorphic learning model termed random graph-based neuro network (RGNN) we develop a joint classification mechanism involving information interaction between multiple RGNNs and realize significant performance improvements in supervised learning for three benchmark tasks, whereby they effectively avoid the adverse impact of the interpretability of NNs on the structure design and engineering practice.}
\end{abstract}


\maketitle

\section{Introduction}
Machine learning and artificial intelligence that rose in this century derive from cognitive science. A variety of neuro networks (NNs) are designed among which deep learning models (DLMs)\upcite{2020Magnetic,ham2019deep,chiu2021predicting,baek2021accurate,cheng2021robust} have been successfully applied to engineering practices such as face recognition, voice recognition, medical data mining. Meanwhile, NNs have innovated the research methodology in many fields owing to their unique properties and outstanding capability in big data environments\upcite{warnat2021swarm,silver2017mastering,havlivcek2019supervised,kavran2021denoising,jia2021unifying,jia2020residual}. Although NNs have brought about innovative changes, their complex internal structures have also led to two grave issues seriously affecting their further developments in other emerging disciplines and research fields, especially the DLMs with massive model parameters and hyperparameters.

The first problem is the real-time performance of NNs. Some DLMs with large width and depth will inevitably contain abundant model parameters. As a result of this, the algorithm complexity problem of DLMs occurs that can make the model training a time-consuming procedure. This shortcoming of NNs directly influences the algorithm efficiency and raises the running cost of those learning models. In addition, time-consuming training can severely obstruct their expansion to some specific fields such as robot control and object detection, etc. To further enhance the practicability of deep learning, scholars have done substantial theoretical and empirical researches.  The most direct and effective way to alleviate the computational burden and accelerate training is to enable a shallow neural network (SNN)\upcite{mignan2019one} for classification and identification. Those schemes provide a potential solution to the above problems since they abandon the complicated connection relationship in DLMs to obtain a quick response in data mining.  
The second problem is the interpretability of DLMs that account for many difficulties to their parameter regulation and structure optimization. It is well-known that the neurons connection patterns are crucial for building a general type of DLMs which reflects that understanding the fundamental principle of DLMs theoretically for big data classification is the key factor affecting the final performance and fully exploring the application potential. Echoing this perspective, NNs in deep learning developed in recent years have reformed their conventional connection mode and more elaborate connectivity patterns gradually occupied the mainstream of NN structure design. The best-known deep learning models like ResNet and MobileNet all have brand-new connection modes. Accordingly, their originative internal connections have made them outperform many traditional NNs.

Advancing this trend, as mentioned above, ResNet\upcite{zhang2018predicting,he2016deep} and MobileNet\upcite{gang2021recognition,sandler2018mobilenetv2} show more powerful abilities in classifying images. Moreover, neural architecture search (NAS)\upcite{xie2019exploring} in deep learning has emerged as a promising direction for optimizing the wiring patterns. The NAS approach usually implicitly relies on a new component termed network generator. This new component contains lots of possible wiring patterns which are sampled from a specific learnable probability distribution. Although some parameters in the network generator are hand-designed, it is obvious that the NAS technique greatly reduced human intervention. These breakthroughs in designing NNs demonstrate a core idea that is contrary to popular belief. Professional knowledge in the structural analysis of NNs has been replaced by NAS. In, a randomly wired deep network is proposed based on several kinds of random graphs which contain Erdős-Rényi (ER), Barabási-Albert (BA) and Watts-Strogatz (WS) model. Besides, these random graph models are converted into directed acyclic graphs (DAGs) to represent signal flow in a deep network. And the experimental results are also perfect. Its performance is better than many fully manually designed learning models and networks generated by NAS methods. Despite the outstanding ability of the method proposed in \upcite{xie2019exploring}, it still perplexes many researchers by its complex structure. And the computation burden makes it difficult to be used in other applications like control, motion planning, and simple data sets classification, etc. It is necessary to improve and simplify the traditional complex NNs with deep architecture.

In this work, we fully exploits the potential application value of random graph theory and develop a new NN model termed RGNN without conventional structure \upcite{mou2021analog,gutig2016spiking,zhang2021self,noe2019boltzmann,kaufmann2020crystal}. In summary, our research findings have three main differences in comparison with those existing works. Firstly, the traditional machine learning models\upcite{luo2021ecnet,hsieh2021automated,lin2018all} always have fixed internal structure, which need users to specify a fixed connection for each neuron and neurons are stacked layer by layer. Yet the most salient feature of RGNN is mainly reflected in weakening the importance of "layer" in the network design and makes the connection mode more flexible. Then, those NNs with multiple layers structure are trained by back propagation algorithm\upcite{robert1989theory} by layers that greatly reduce their training efficiency and because of that, corresponding operating costs will also increase significantly. However, we find that replacing Exponential Moving Average (EMA) based ADMM optimizer with traditional ADMM\upcite{shi2014linear,shen2012distributed,wang2019global} can improve the test accuracy, with enhanced robustness of the model. In addition, EMA-ADMM markedly reduces the computational cost without affecting the accuracy compared to back propagation. Finally, mainstream NNs directly draw final output based on their own classification results. Unlike this, we can generate lots of RGNNs with different architectures, using random graph algorithm with different parameters. Furthermore, we introduce a joint classification mechanism, whereby enables mutual communication between every single RGNN with different structures which can make up for the deficiency of itself, thus we find that the multiple RGNNs scheme under joint decision making mechanism can obtain significantly higher accuracy in supervised learning tasks compared to the average level of the single RGNN and other state-of-the-art NN algorithms.

\begin{figure*}[t]%
	\centering
	\includegraphics[width=.95\textwidth]{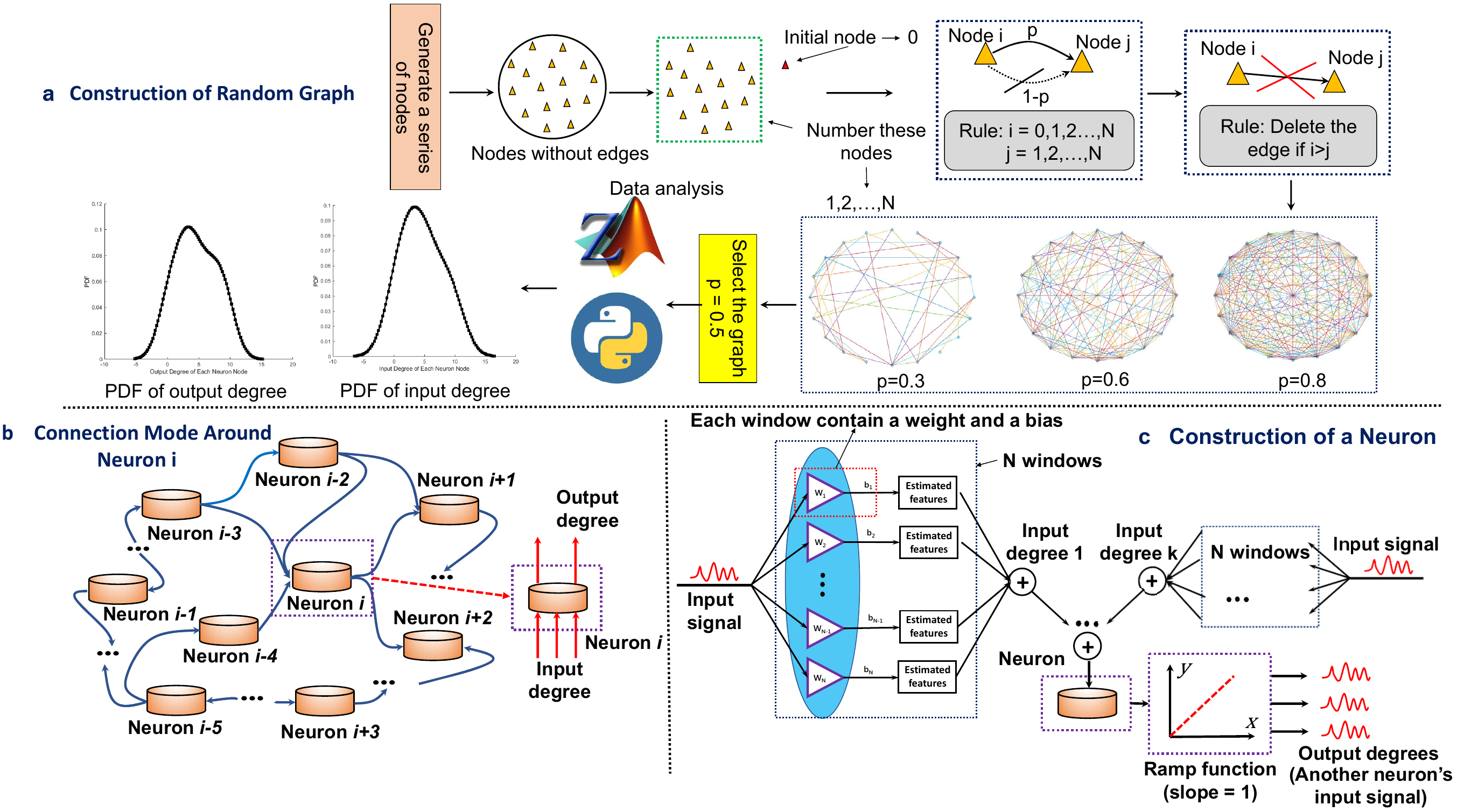}
	\caption{\emph{\footnotesize{Overview of RGNN I: \textbf{a}.The connection pattern of the graph depends on the connection probability $p$. Users only need to designate a parameter $p$ for the graph so that it can be self-generated. The rule that is written as “delete the edge if $i>j$” is set for avoiding self-loops. And it means that the signals can not “return” in the NN model. Moreover, we can use MATLAB or Python tools to analyze the graphs and their corresponding data. Here we show the estimated probability density function image of the output degree and input degree in generated graphs. \textbf{b}. The detailed illustration of the connection mode is shown here. The neuron's output degree represents the number of neurons pointing to other neurons. Meanwhile, a neuron's input degree is defined as the number of other neurons pointing to the neuron. \textbf{c}. The neuron is composed of its input degrees and each input degree consists of several mapping windows.
	}}}\label{fig_1}
\end{figure*}
\begin{figure*}[htbp]%
	\centering
	\includegraphics[width=.95\textwidth]{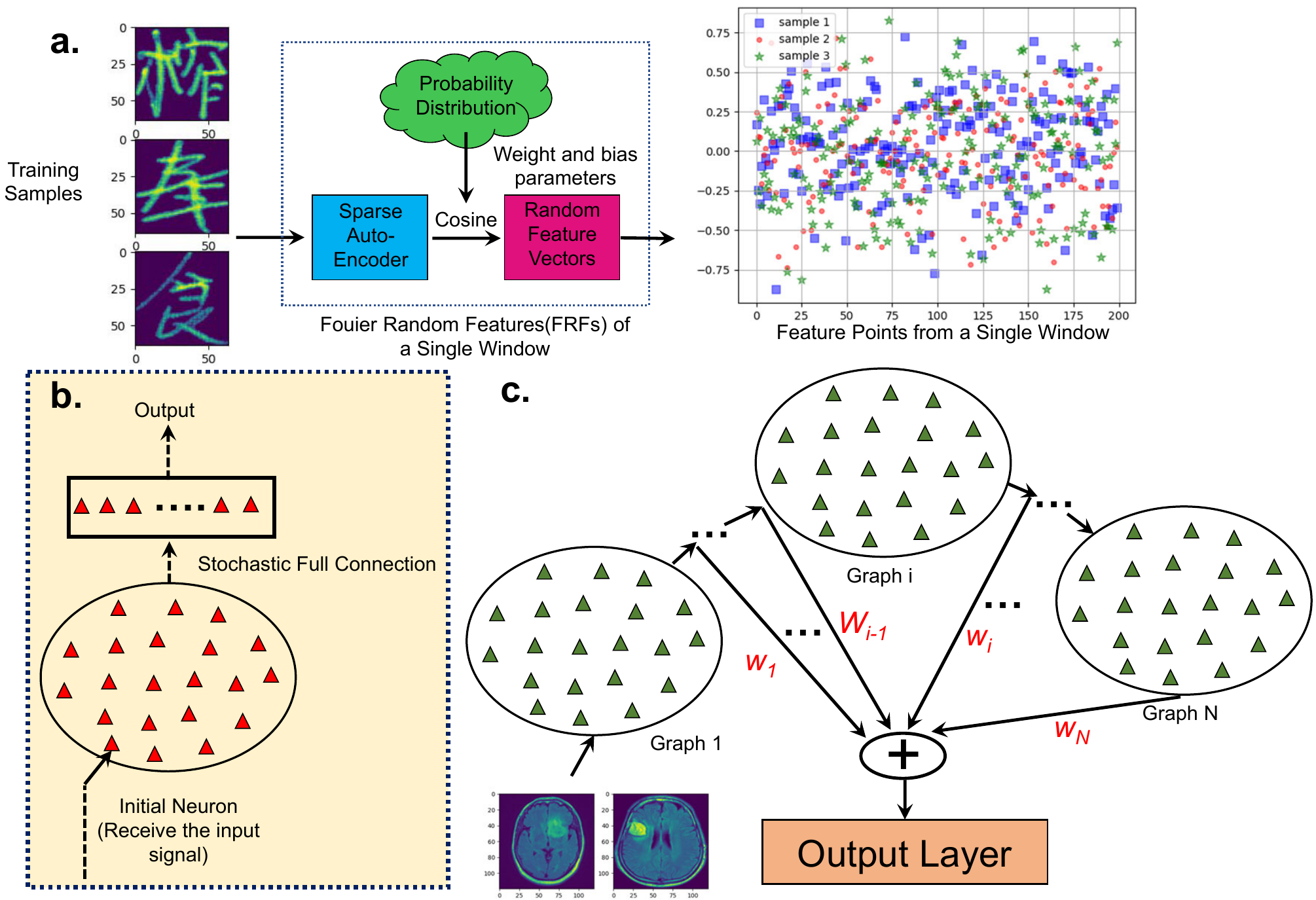}
	\caption{\emph{\footnotesize{Overview of RGNN II: \textbf{a}.The neurotic window is the basic unit of a single neuron in the random graph. The first annulus in the neurotic window is a sparse auto-encoder that obtain the preliminary sparse representation of the input. Next, the FRFs method is adopted to implement random features extraction with weight and bias sampled from a specific probability distribution. In this process, we use the cosine function to replace the $e^{j\omega^Tx}$ for avoiding the usage of imaginary number. Via these operations, each window can calculate a series of feature points that are crucial to the composition of input degree. \textbf{b}. The neurons in a single graph are established according to its internal connection relationship that are independent of other graphs. \textbf{c}. The entire RGNN model contains several random graphs and each graph's output represent feature information of different levels. The output of the previous random graph is the input of the next adjacent random graph. And the output of all the random graphs is mapped to the label space by connection weight obtained by training}}}
	\label{fig_2}
\end{figure*}

\section{Results}
\subsection{RGNN: a stochastic neuromorphic learning approach with layer-weaken structure}

SNNs are quite different from deep learning models like convolution neural networks (CNN, such as ResNet-18 and ResNet-50) or deep belief network in structure. The most obvious characteristic of SNNs in comparison with CNN is that they only contain a visible layer and a hidden layer connected to output layer. The multi-layer structures and back propagation (BP) algorithm for each layer are abandoned in SNN. They are usually constructed by only one kind of activation functions for feature mapping. On this basis, it is no longer necessary to use BP algorithm for derivation layer by layer that economizes a lot of time. Based on these successful practices of the NNs without deep architecture, we establish a different neuro model by random graph theory.

The developed RGNN has the capability of generating substantial neurons with a random connection, namely, can be constructed without much professional knowledge, only by specifying the connection probability of each neuron. For a single random graph, we make overviews of its connection mode and calculation process (Fig.~\ref{fig_1}, Fig.~\ref{fig_2}). The generation process of the connection relations of a single random graph is simple and not occupying too much computing resources (Fig.~\ref{fig_1}\textbf{a}). By and large, it can be divided into three stages. In the first phase, we establish a series of neurons according to the original setting. Second, we need to specify the initial neuron and number them. Finally, those neurons are connected by edges that follow a certain connection relationship established by probability $p$. The detailed illustration of the connection mode of a specific neuron in the random graph is shown in Overview of RGNN I (Fig.~\ref{fig_1}\textbf{b}). And it shows that each neuron consists of many input degrees calculated by other neurons' output. The output signal will flow into other neurons that the neuron points as input degree (Fig.~\ref{fig_1}\textbf{c}). The basic unit of each input degree is the mapping window. Moreover, we introduce a Fourier Random Features (FRFs) method to calculate each window in neurons (see in Fig.~\ref{fig_2}).  We make a brief description of FRFs in extracting the feature points from Chinese characters (see in Fig.~\ref{fig_2}\textbf{a}). The neurons in a single graph are only corresponding to its neighbor neurons. After all these neurons in the graph are calculated, we make a stochastic full connection to them and use the connection matrix as the output of a single random graph (see in Fig.~\ref{fig_2}\textbf{b}). Finally, several random neuron graphs are established and they are all connected to the output layer to (see in Fig.~\ref{fig_2}\textbf{c}). Particularly, the introduced RGNN abandon the conventional hierarchical structures of which feature mapping operations are carried out layer by layer. The neurons in each graphs are randomly connected to each other, which means that there are no obvious hierarchical structures between neurons in the random graphs. Furthermore, the final output of each random graph in the network is directly mapped to the output layer. Based on the above characteristics, that is to say, the proposed RGNN weakens the significance of "layer" in structural design, thereby realizing the diminishment of manual participation.

It seems that the RGNNs computational process in practical applications will be cumbersome and directly affect its property of efficiency. Accordingly, we use a laptop equipped with an AMD Ryzen 9-4900H with RTX 2060 to test the efficiency of the RGNN's construction process. In the experimental settings, the RGNN contains three neuron graphs and each graph restrains 16, 18, 20 neurons with 50 feature points per neuron. The NORB\upcite{lecun2004learning}, MNIST\upcite{lecun1998gradient}, Pima Indians Diabetes (PID)\upcite{smith1988using}, Wisconsin Breast Cancer (WBC)\upcite{WBC}, Fashion MNIST\upcite{xiao2017fashion}, UMIST\upcite{umist}, Extended YaleB\upcite{lee2005acquiring}, ORL data set\upcite{ORL} are used in this efficiency test to establish neuron nodes for random graphs. After getting the accuracy of RGNN and the time spent establishing the network (The time spent from the start of the algorithm to the completion of all nodes or layers calculation), we set the structure of other comparison algorithms according to its accuracy to ensure that the accuracy of various methods is roughly consistent. The batchsize keeps the same as the number of training samples. We organized the experimental results into a table (see in Supplementary Information, \emph{Table S1}). It is only natural that the construction time of an RGNN also depends on the data dimension of each node largely. On balance, the proposed RGNN structure is efficient and does not occupy too many computing resources allocated to the training optimization algorithm.

\subsection{Two Applications of a Single RGNN}

We conduct two classification experiments on face attribute recognition data (CelebA data) and Fashion MNIST data using a single RGNN to assess the first potential limitation and preponderance by comparing it with many state-of-the-art NN models from the perspective of error rate, parameter quantity, construction time, training time and testing time, etc. The face attribute recognition is a binary classification problem that can test the corresponding capability of single RGNN. Meanwhile, the Fashion MNIST is used for multi-class classification.
\subsubsection{\emph{Face Attribute Recognition (Binary Classification)}}

\begin{figure*}[htbp]%
	\centering
	\includegraphics[width=.9\textwidth]{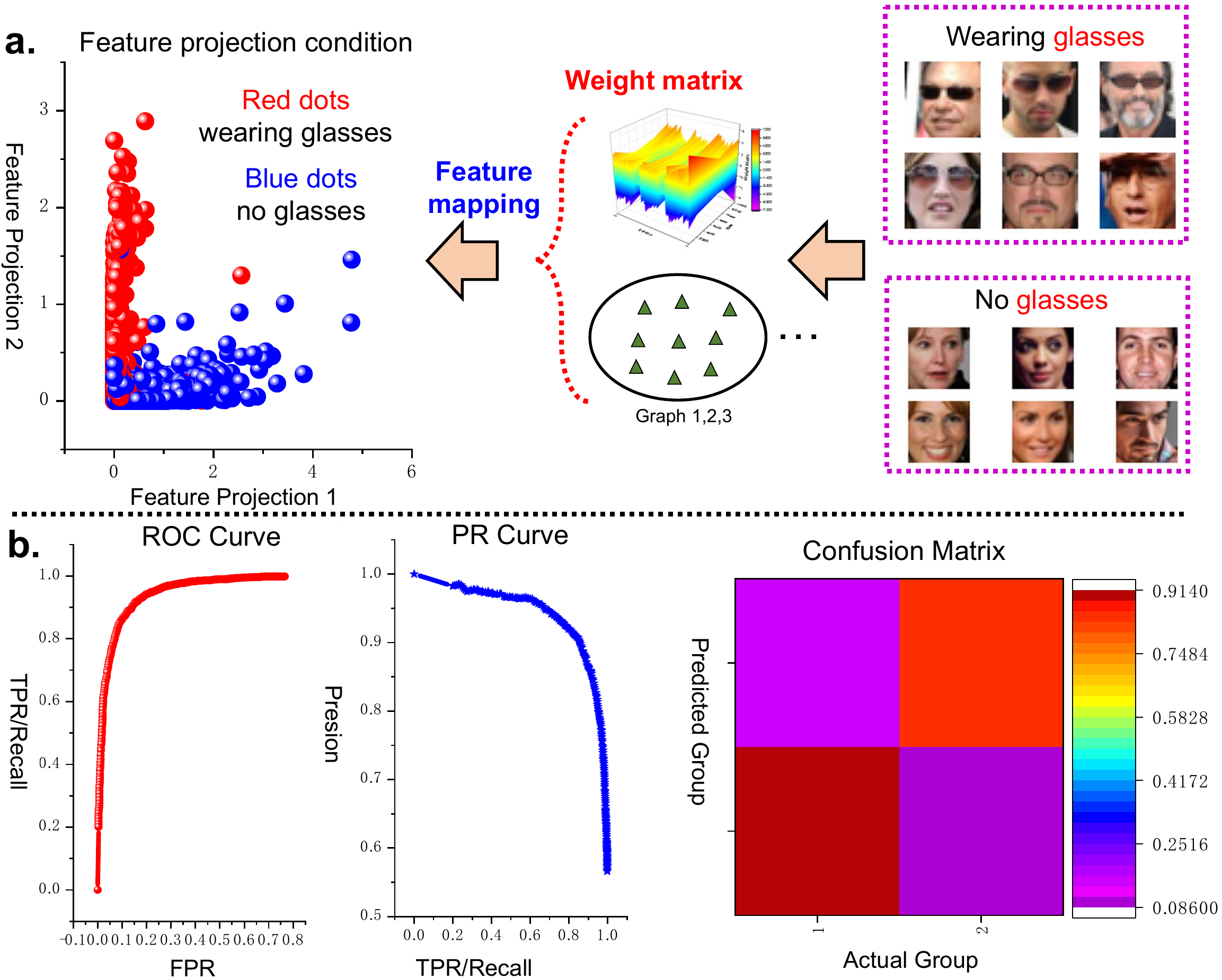}
	\caption{\emph{\footnotesize{Face attribute recognition results using CelebA data: \textbf{a}. This classification experiment is designed for verifying the binary classification capability of a single RGNN. We divide face images in CelebA dataset into two categories : wearing glasses and not wearing glasses. The feature projection condition can show that single RGNN possesses good capability of binary classification. \textbf{b}. Another visualization of experimental results is obtained after the quantitative analysis of the experimental data. The first is the ROC curve of which longitudinal coordinates stand for recall/true positive rate (TPR) and horizontal ordinates denote false positive rate (FPR). It can show the recognition accuracy at a certain threshold. In the middle is the PR curve that is calculated by precision (longitudinal coordinate) and recall/TPR (horizontal ordinate). The most right-hand side is the confusion matrix of this binary classification experiment. It can be concluded that the overall accuracy of RGNN in glasses attribute recognition attains more than 90\%.}}}\label{fig_3}
\end{figure*}

\begin{figure*}[htbp]%
	\centering
	\includegraphics[width=.95\textwidth]{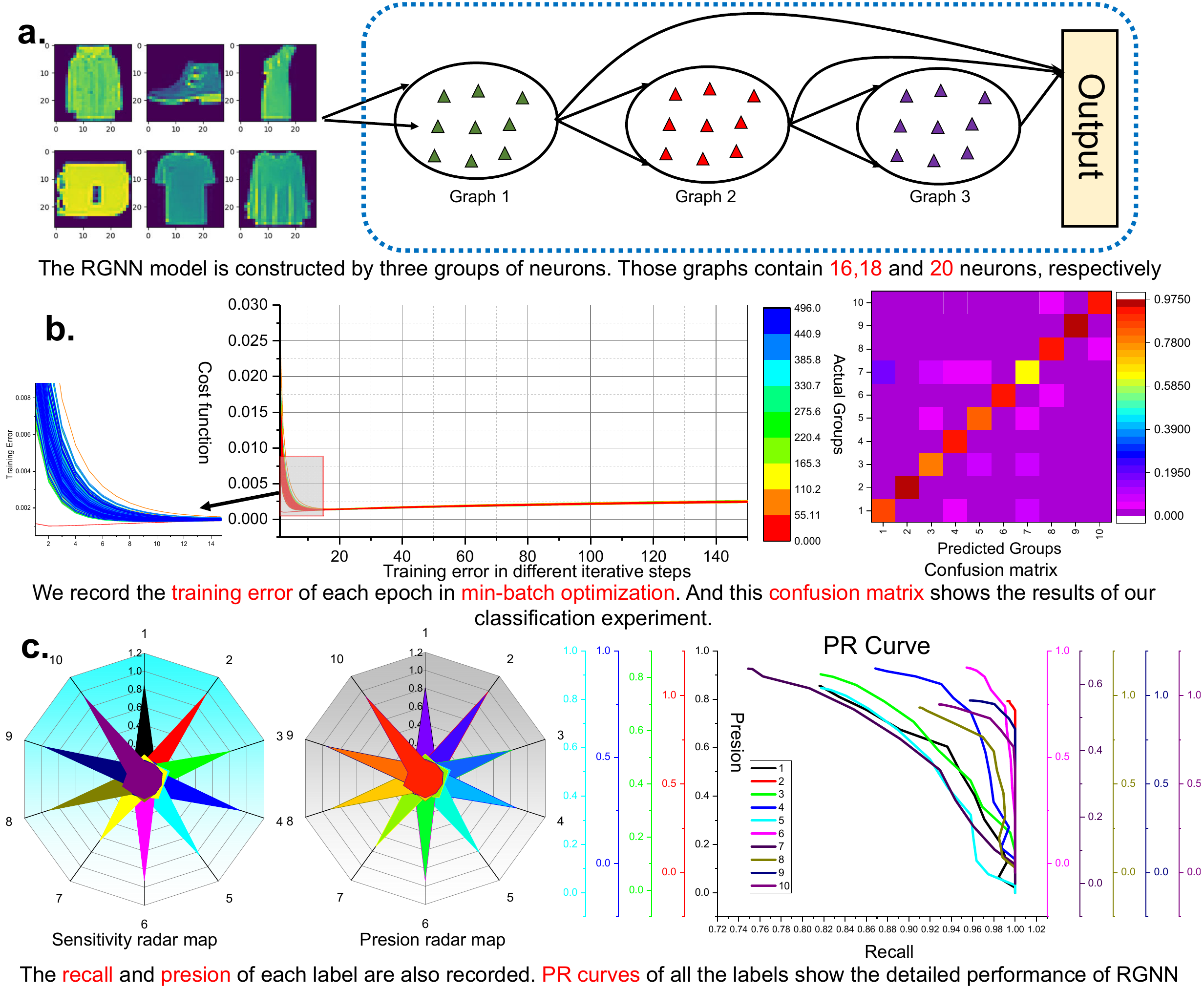}
	\caption{\emph{\footnotesize{Universal Approximation Capability of RGNN on Fashion MNIST data. a. The RGNN designed in this experiment contains three random graphs which have 20, 22 and 24 neurons, respectively. The connection probability of neurons in each graph is set to 0.5. b. We use the min-batch optimization in the training phase of RGNN. The left figure is the cost function we have defined in each epoch during training. It illustrate that the EMA-ADMM optimizer combined with the idea of min-batch training can realize the optimization of cost function in less iteration steps of each training epoch. The right figure shows the confusion matrix in this multi-class classification experiment. We can find that in addition to the samples represented by label 7 (shirt), almost every category can be identified correctly with a probability of nearly 90\% or even more than 90\%.It is considered that the shirts are very similar in appearance to some T-shirt (Label 1), resulting in many shirts being misclassified as T-shirts. c. We calculate the sensitivity and precision of each label, meanwhile those evaluation indexes are transformed into radar map and PR curve.}}}\label{fig_4}
\end{figure*}
We will primarily test the binary classification capability of RGNN through the well-known face database termed CelebA\upcite{liu2015deep}. CelebA dataset is provided by the Chinese University of Hong Kong and has been widely used in face-related computer vision training tasks for a long time. Since this classification experiment is designed for verifying the binary classification capability of a single RGNN, we divide face images in CelebA dataset into two categories : wearing glasses and not wearing glasses. Besides, we select glasses as the facial attribute that contain sunglasses, myopia glasses, swimming goggles, etc. The wide variety of glasses in dataset also bring great difficulties to the classification. The RGNN in this binary classification experiment is made up of three neuron graphs. The three graphs contains 16, 18, 20 neurons, and each neuron has 50 feature points. The connection probability $p$ is set to 0.5. We run the algorithm of RGNN using DELL PRECISION M4800 which is equipped with Intel Core i7-4910MQ CPU. 

\noindent\textbf{Finding 1: }\emph{The single RGNN can complete low time-consuming big data binary classification tasks without affecting its accuracy.}

A formal statement of this finding and a detailed discussion are given in the following.

In this experiment, we specially conduct a feature projection operation for the testing results of RGNN classifier in which we have marked different kinds of samples at distinct points with various colors. And it is easy to distinguish which sample points belong to the same class from multitudinous feature points (Fig.~\ref{fig_3}\textbf{a}). We mainly visualize the confusion matrix, PR curve and ROC curve of the classifier to quantify the testing results (Fig.~\ref{fig_3}\textbf{b}).

These visualization of experimental results is obtained after the quantitative analysis of the experimental data. The first is the ROC curve of which longitudinal coordinates stand for recall/true positive rate (TPR) and horizontal ordinates denote false positive rate (FPR). The PR curve is calculated by precision (longitudinal coordinate) and recall/TPR (horizontal ordinate) (Fig.~\ref{fig_3}\textbf{b} ). The PR curve and ROC curve of RGNN classifier (Fig.~\ref{fig_3}\textbf{b}) show that our RGNN classifier has a superb performance in the facial attribute recognition. The most right-hand side is the confusion matrix of this binary classification experiment from which many evaluation indexes like the final TPR, FPR and precision can be concluded. The positive samples in the experiment refer to the samples containing a subject wearing glasses. And TP, referring to true positive, is equal to the number of positive samples correctly identified as positive samples. On the contrary, negative samples denote the samples containing a subject without glasses. Accordingly, in terms of literal meaning, TN means that the NN doesn't predict negative samples as positive samples. Thus, we use TN to represent the number of negative samples identified as negative samples. Besides, FN stands for the number of positive samples predicted as negative samples; FP represents the number of negative samples predicted as positive samples. The overall accuracy (OA) is calculated as OA=(TP+TN)/(TP+TN + FP+FN) which can reach 90.25\% under this database. Meanwhile, OA for each group can also be found in visual results of the 3D confusion (see in supplementary information). Besides, we conduct comparative experiments using many state-of-the-art machine learning algorithms for RGNN to emphasize its superior performance (see in Supplementary Information, \emph{Table S2}).

\subsubsection{\emph{Fashion MNIST Classification (Multi-Class Classification)}}

To enhance the universality of the single RGNN application results, multi-class classification is an important test to show the prediction capability dealing with multi-categories complicated data. Fashion MNIST is a typical dataset with ten categories which is provided by the research department of Zalando, a German fashion technology company. The size, format and training set/testing set partition of Fashion MNIST are completely consistent with the original MNIST. In detail, this database contains 1) T-shirt, 2) trousers, 3) pullovers, 4) dresses, 5) overcoats, 6) sandals,7) shirts, 8) sneakers, 9) handbags and 10) short boots.

\noindent\textbf{Finding 2: }\emph{The single RGNN can complete low time-consuming big data multi-class classification tasks without affecting its accuracy.}

The  formal statement of this finding and a detailed discussion of corresponding visualization results are given in the following.

Since its complexity is greater than that of binary data, we increase the number of neurons in each random graph in comparison with the previous binary classification test. They contain 20, 22 and 24 neurons, respectively. The dimension of each neuron is consistent with the setting in the binary classification experiment. Resembling the visualization in Fig. \ref{fig_3}, the results of this multi-class classification experiments are also presented to readers in various visual ways (Fig. \ref{fig_4}).
We show the architecture of RGNN used in this classification (Fig. \ref{fig_4}(\textbf{a})) that uses three graphs to characterize the features of data.  The validity of the proposed EMA-ADMM optimizer in training the RGNN is confirmed and the confusion matrix that displays the recalls of each category of samples indicate that RGNN has excellent multi category classification ability (Fig. \ref{fig_4}(\textbf{b})). Furthermore, we also give two radar map graphs corresponding to the precision and sensitivity of each type of sample (Fig. \ref{fig_4}(\textbf{c})) that can make the classification results more intuitive. Accordingly, followed by the establishment of two radar map graphs is the analysis of precision and sensitivity by setting different thresholds of each type of sample. Similar to the binary classification experiment, we conduct comparative experiments using many state-of-the-art machine learning algorithms and testing accuracy, time and other information are displayed in the form of tables (see in Supplementary Information, \emph{Table S3}).
\begin{figure*}[htbp]%
	\centering
	\includegraphics[width=1\textwidth]{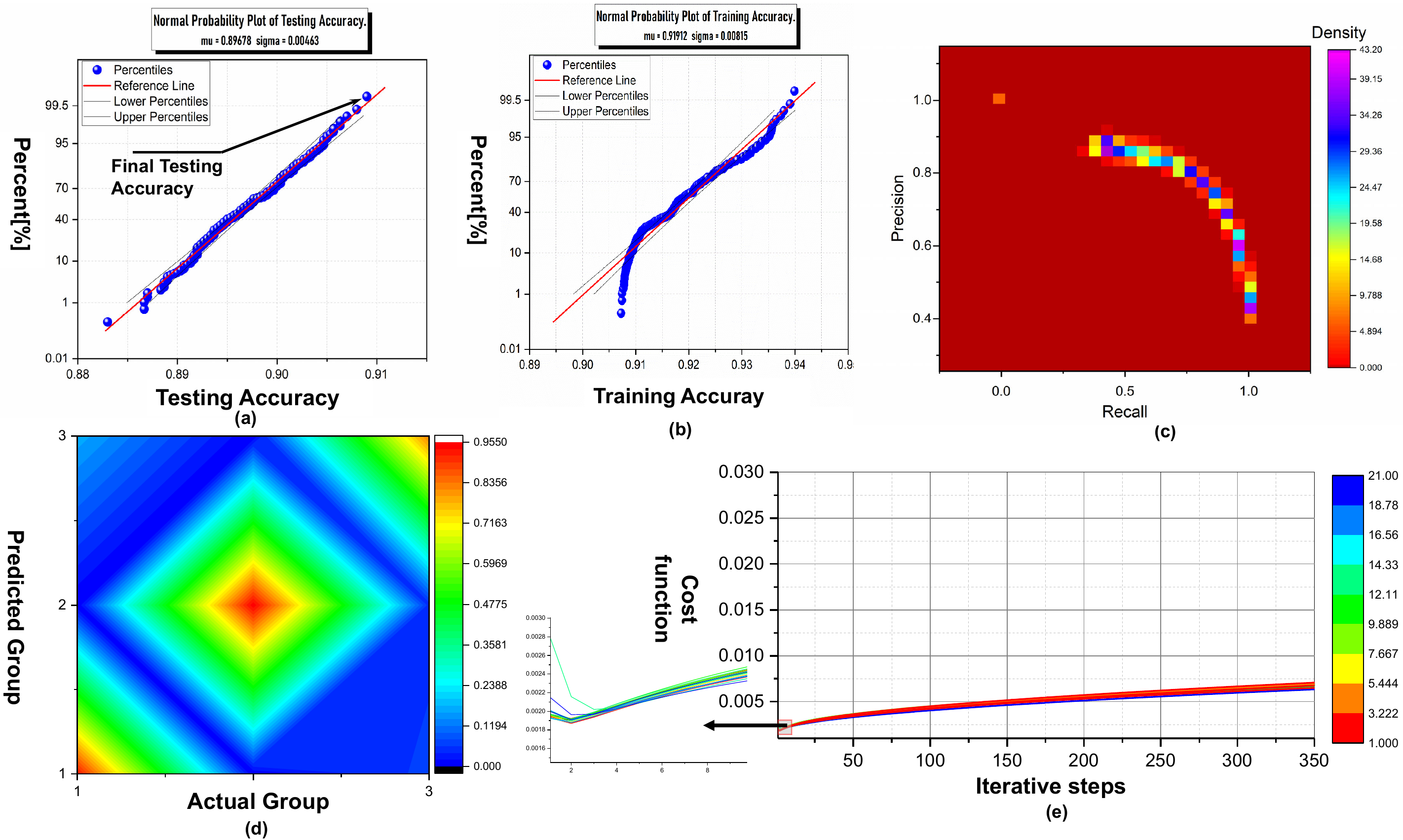}
	\caption{\emph{\footnotesize{Experimental Results for M-RGNNs on Lung Cancer: (1). The two graphs (a) and (b) show the training and testing performance of the M-RGNNs. It can be concluded that different connection probability in rewiring neurons can affect the training and testing nature of RGNN. With different connection architecture, the training accuracies vary from 90.5\% to 94\%. Meanwhile the testing accuracies in identifying unknown histopathological images vary from 88.3\% to 90.9\%. Under our designed inference mechanism, the final testing accuracy can reach about 90.83\% which outperforms 99\% of the performance of a single RGNN with fixed connection probability. (2). The graphs (c) and (d) are visualizations of the PR curve in density form and the final confusion matrix in isoline form. Those visualization results, whether in 2D form or 3D form, can directly show that a preliminary diagnosis can be made with very low error rates using multiple RGNNs. (3). Subfigure (e) reveals the performance of EMA-ADMM optimizers in M-RGNNs and the results have shown that the proposed optimizer still can maintain the original validity in dealing with microscopic image of cell tissue.}}}\label{fig_5}
\end{figure*}

\subsection{Classification and Diagnosis of Cancer by Multiple RGNNs}

On the basis of the excellent performance of the single RGNN we have stated, we utilize the random graph algorithm with different connection probability $p$ to generate multiple RGNNs (M-RGNNs) and apply the developed joint classification mechanism into practice. Thus the users can completely assign the task of network structure optimization to random graphs.

\noindent\textbf{Finding 3: }\emph{Multiple RGNNs based neuromorphic learning with diversified structures has an autonomous near-optimal architecture exploration (ANoAE) capability, which can elevate the accuracy of artificial neural networks (ANNs) in supervised learning and markedly reduce users' participation. }

The dataset\upcite{Cancer} used in this test contains 25,000 histopathological images with 5 classes. All images are 768$\times$768 pixels in size and are in jpeg file format. The images were generated from an original sample of HIPAA compliant and validated sources, consisting of 750 total images of lung tissue (250 benign lung tissue, 250 lung adenocarcinomas, and 250 lung squamous cell carcinomas) and 500 total images of colon tissue (250 benign colon tissue and 250 colon adenocarcinomas) and augmented to 25000 using the Augmentor package. In general, there are five classes in the dataset, each with 5,000 images, being: 1) Lung benign tissue, 2) Lung adenocarcinoma, 3) Lung squamous cell carcinoma, 4) Colon adenocarcinoma, 5) Colon benign tissue. In our experimental settings, we divide this dataset into two parts according to different lesions. Thus we can discuss these two situation individually. The M-RGNNs tested in these experiments is made up of 50, 100, 150, 200, 250, 300 RGNNs that are generated by our introduced random graph algorithm. Each RGNN contains three graphs and those graphs internally include 20, 22 and 24 neurons, respectively. Different to the previous experiments, we set the number of feature points contained in each neuron as 60. To ensure structural diversity of those multiple RGNNs as much as possible, we do not duplicate connection probability $p$ corresponding to each RGNN. Accordingly, we design a random number generator subject to a specific probability distribution to obtain the connection probability without any human intervention. Ultimately, we can derive the final classification results of those pathological tissue section images according to the principle of the minority is subordinate to the majority considering all the prediction results of the M-RGNNs. The details of the joint classification algorithms in M-RGNNs can be found in \emph{Method}.

The first application is the diagnosis of lung cancer. While verifying the ANoAE capability of M-RGNNs, we have used a variety of mainstream NNs for comparative experiments. The parameters of these NNs and the detailed classification accuracy of them can be found in Supplementary Information, \emph{Table S5}. We selected M-RGNNs with 300 subnets to visualize the experimental results. Notwithstanding some RGNNs with low accuracy in M-RGNNs, the final accuracy of M-RGNNs still exceeds 99.5\% of RGNNs, as well as the average accuracy of all the RGNNs (Fig. \ref{fig_5}(\textbf{a}), Fig. \ref{fig_5}(\textbf{b})). The PR curve and confusion matrix illustrate the effectiveness of M-RGNNs from another perspective (Fig. \ref{fig_5}(\textbf{c}), Fig. \ref{fig_5}(\textbf{d})). The EMA-ADMM optimizer dealing with cancer images can still maintain a fast convergence rate (Fig. \ref{fig_5}(\textbf{e})). Based on these experimental results, it is obvious that our M-RGNNs are beneficial to improving the performance of a single RGNN without the knowledge of the impact of internal connection structure on network performance. That is to say, we don't need human intervention in the internal structure of each RGNN. Through its self-generation mechanism and the proposed combinative diagnosis mechanism, we can also obtain results that are better than 99\% of the performance of a single RGNN. It is equivalent to that the developed M-RGNNs can find the approximate optimal structure of RGNN independently without complex theoretical analysis.

For another kind of cancerous image taken from the colon, we also use the developed M-RGNNs to realize a combinative diagnosis. The basic parameters of neurons keep consistent with the lung cancer diagnosis we have conducted. These detailed experimental results are listed in Table. S4 (see in Supplementary Information).

\begin{figure*}[htbp]%
	\centering
	\includegraphics[width=1\textwidth]{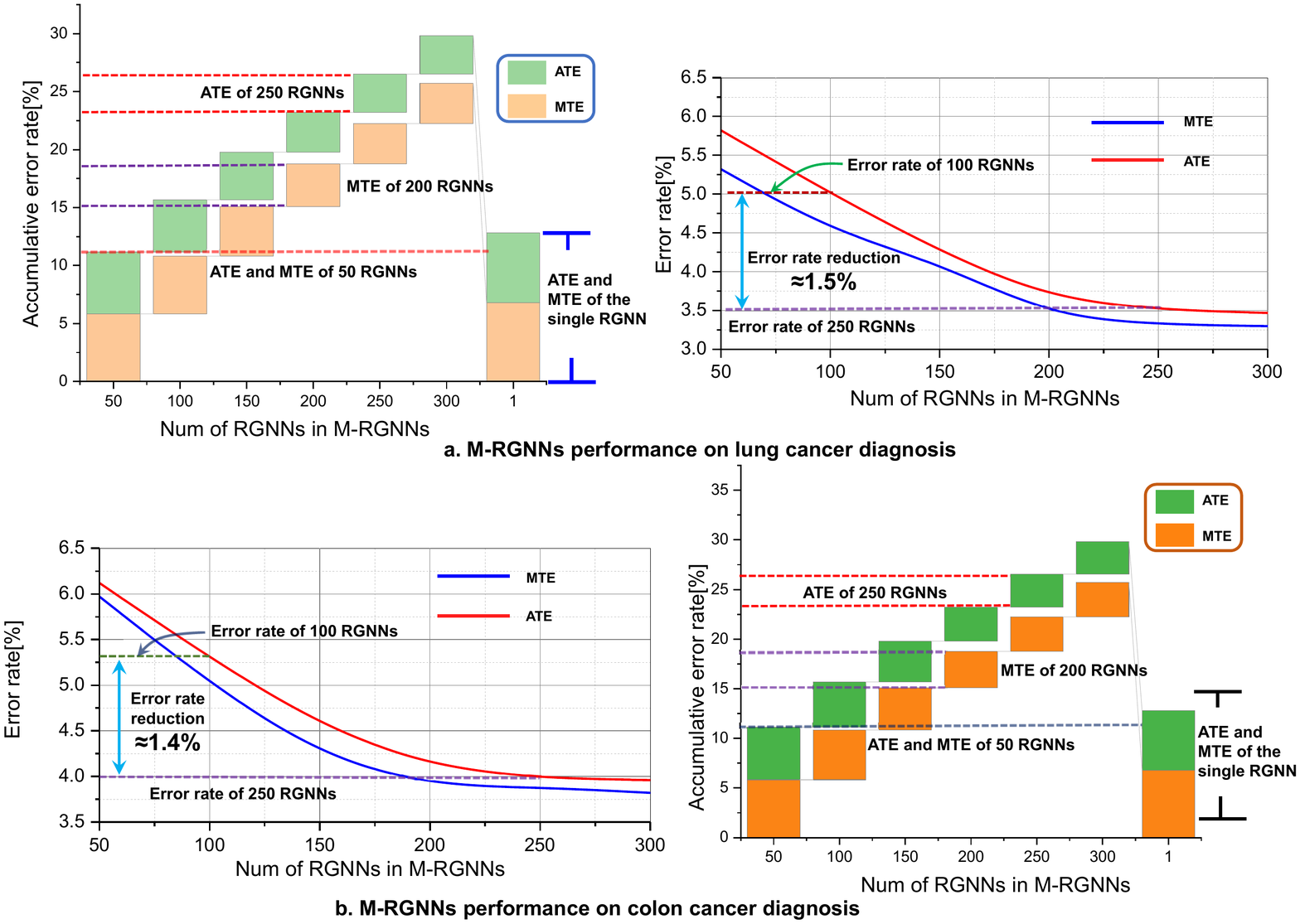}
	\caption{\emph{\footnotesize{\textbf{ATE and MTE analysis of M-RGNNs on lung cancer benchmark and colon cancer benchmark.} Performance of M-RGNNs for disease diagnosis using lung cancer database: (a) The ATE and MTE of M-RGNNs decrease with the increase of the number of RGNN in M-RGNNs. We can find that the ATE under M-RGNNs-250 is 1.5\% lower than that under M-RGNNs-100. This result shows that M-RGNNs with sufficient RGNNs can diminish the ATE and MTE in comparison with the single RGNN. Performance of M-RGNNs for disease diagnosis using colon cancer database (b) is presented in the same manner as that in (a).}}}\label{fig_6}
\end{figure*}

\section{Discussion}
Driven by the need to solve the problems of high operating cost and interpretability, we apply the RG theory into NN design and develop an approach using multiple RGNNs with different inner connection mode that abandon the conservative deep structures and realize significant reductions of structural parameters. It fully utilizes the diversity of the designed random graphs to obtain relatively better prediction results without the human prevention which is equivalent to the capability of autonomous near-optimal architecture exploration (ANoAE). 

We primarily confirm the effectiveness of the single RGNN through two famous benchmark tasks, involving binary classification and multi-class classification in big data environment. Aiming at providing ideas in dealing with time-consuming problems, the developed RGNN without complex structure and massive optimization parameters, shows an obvious decline in constructing its neurons and training compared to deep learning without affecting the accuracy (Supplementary Information, \emph{Table. S1}, \emph{Table. S2}, \emph{Table. S3}). We further collect the minimum test error (MTE) and the average test error (ATE) shown in percentage (Supplementary Information, \emph{Table. S2}, \emph{Table. S3}) which is conducive to quantitative analysis of testing. The ATEs of RGNN are lower than other approaches in experiments, therefore it possess the best generalization performance (accuracy on test data). When the time consumption of RGNN is similar to that of other mainstream deep learning techniques, we can train a classifier with better generalization ability using RGNN framework, and this superiority is not possessed by deep NNs. 

The ANoAE capability of M-RGNNs brought by several RGNNs as well as their joint classification mechanism is the centrepiece we demonstrate in this work. We specially design biologically experiments in which cancer histopathological images benchmark data are used for training and validation. Taking the diagnosis of colon cancer as an example, all the details of the diagnosis and its corresponding results like PR curve, ROC curve (see in Supplementary Information, \emph{Fig. S1} and \emph{Fig. S2}) are similar to the diagnosis results of lung cancer (We also selected M-RGNNs with 300 subnets to visualize the experimental results.). In the meanwhile, the testing accuracies in identifying colon cancer vary from 83.46\% to 88.84\%, among which the median accuracy only reaches about 85\%, while that of M-RGNNs under our designed inference mechanism can reach about 90.7\%. This accuracy is about seven percentage points higher than the RGNN with minimal accuracy in M-RGNNs (see in Supplementary Information, \emph{Fig. S4}). Furthermore, the final accuracy is also four percentage points higher than the average accuracy of the 300 RGNNs in M-RGNNs (see in Supplementary Information, \emph{Fig. S3}), also obviously exceed the RGNN with maximum accuracy in M-RGNNs by about 2 percentage points (see in Supplementary Information, \emph{Fig. S3}), that is, it demonstrates that the application scope of M-RGNNs based combinative diagnosis can be extended to medical pathological image diagnosis, by taking advantage of our layer-weaken RGNNs with a specially designed classification mechanism. The ATE and MTE analysis of M-RGNNs in histopathological images benchmark task (Fig. \ref{fig_6}) validate the conclusion stated in \emph{Finding 3}. With the participation of enough RGNNs, the accuracy and error rate of M-RGNNs in these two benchmark task can be improved, without using much professional knowledge of ANN. More importantly, the finding of ANoAE capability alleviates the limitation of NN models brought by their interpretability. Such mutual communications between multiple NNs have a bright application prospect in the future.

\section{Materials and Methods}

\subsection{Random Graphs Theory}
The traditional NN models are constructed mainly by users themselves, that are entirely different from our scheme. Rich professional knowledge of designers is of key importance in improving the NN models in previous works. By contrast, the complexity of designing random graph is much lower than that of designing a classification and recognition network with excellent performance. How to establish a random graph steering the wired mode of internal structure takes critical role in surpassing other data mining models in performance. The detailed algorithm of it is shown in supplementary information, \emph{Fig. S5}.

More specifically, we use $\otimes$ to represent the “straighten” operation which means that the nodes in the random graph are spliced horizontally according to the its number. The random topology graph can be denoted as $\mathcal{G} = \left(\mathcal{V},\mathcal{E}\right)$. $\mathcal{V} = \left\{\nu_1,\nu_2,\cdots,\nu_n\right\}$ is the set of the nodes in the graph. $\mathscr{A}\left(\nu_i\right)$ stand for the set of the input nodes of $\nu_i$ . $\mathscr{O}\left(\nu_i\right)$ is denoted as the set of the output nodes of $\nu_i$. $\mathcal{N}_i$ represent the set of indexes of the nodes in $\mathscr{A}\left(\nu_i\right)$. $\mathcal{M}_i$ denote the set of indexes of the nodes in $\mathscr{O}\left(\nu_i\right)$. $\mathcal{P}$ stand for the set of indexes of $\mathcal{V}$. And $\mathcal{P}\setminus\mathcal{N}_i$ represent the indexes set $\left\{i|i\in\mathcal{P}, i\notin\mathcal{N}_i\right\}$.

\subsection{Computation of Neurons}

Random graphs and nodes detaching from actual background are meaningless, which indicates that we must endow actual meanings of nodes in the graph. Accordingly, we use neurons that represent the potential features in NMs to make sense of nodes in random graphs.

A good feature representation of the input is necessary for realizing an outstanding performance. Naturally, we employ the sparse auto-encoder (SAE) that could be counted as a crucial tool to complete feature representation, driving a set of sparse and compact features. The original training samples are denoted as ${\textbf{X}}\subseteq\mathbb{R}^{N\times N_f}$ that contain $N$ samples and each sample has $N_f$ feature points. Unsupervised training of SAE can be formulated in the following way:
\begin{align}\label{eq:CN1}
\begin{split}
\arg\min\limits_{\textbf{W}}||\textbf{Z}\textbf{W} - \textbf{X}||^{\sigma_1}_v + ||\textbf{W}||^{\sigma_2}_u
\end{split}
\end{align}
where $\textbf{Z} = \textbf{X}\textbf{W}_s$ represent the random features of the input and parameter $\textbf{W}_s$ stand for weight matrices sampled from a specific probability distribution.

We can set $\sigma_2 = u = 2$, $\sigma_1 = v = 1$ and this optimization can be solved in our proposed way using EMA-ADMM optimizer which will be illustrated in the next section. We can obtain the solution of the optimization problem defined above which can be denoted as $\textbf{W}^*$. Hence the output of the SAE can be defined as $\textbf{F} = \textbf{X}\left(\textbf{W}^*\right)^{\mathrm{T}}$.

Then the output of the SAE $\textbf{F}\subseteq\mathbb{R}^{N\times N_f}$ is defined as the input signal of graphs. In particular, signal $\textbf{F}$ can be subdivided into $\left[{\textbf{f}}^{\mathrm{T}}_1,\cdots,{\textbf{f}}^{\mathrm{T}}_{N_s}\right]$. On this basis, the Fourier kernel approximation
is applied for initial neuron containing the random features. The $d$-dimension approximation for initial neuron can be formulated as follows.
\begin{align}\label{eq:CN2}
\begin{split}
&\mathscr{F}_l\left({\textbf{f}}_i\right) = \frac{1}{\sqrt{d}}\left(\sqrt{2}\cos\left(\omega^{\mathrm{T}}_{1,l}{\textbf{f}}_i + b_{1,l}\right),\cdots,\sqrt{2}\cos\left(\omega^{\mathrm{T}}_{d,l}{\textbf{f}}_i + b_{d,l}\right)\right)
\end{split}
\end{align}
where the weights $\omega^{\mathrm{T}}_{i,l}$, $b_{i,l}, i = 1,2,\cdots,d$ are sampled from a given distributions $\mathscr{P}\left(\omega\right)$ and $U\left[0,2\pi\right]$, respectively.

Finally,  a FRF of initial neuron can be denoted as follows.
\begin{align}\label{eq:CN3}
\begin{split}
\mathcal{F}_l = \left(\begin{array}{c}
\mathscr{F}_l\left(\textbf{f}_1\right)\\
\cdots\\
\mathscr{F}_l\left(\textbf{f}_N\right)\\
\end{array}\right)\subseteq\mathbb{R}^{N\times d}
\end{split}
\end{align}

Meanwhile, a single FRF can be seen as a window constructing a complete neuron.

Usually a neuron is established with ample feature information, thus we can calculate many other FRFs to enrich the neuron. The composite FRFs of initial neuron is formulated as $\mathcal{F} = \left[\mathcal{F}_1,\mathcal{F}_2,\cdots,\mathcal{F}_M\right]$. And we can finally obtain the final output of the initial neuron $\mathscr{N}_1 = \varphi\left(\mathcal{F}W_{n_1} + \beta_{n_1}\right)\subseteq\mathbb{R}^{N\times dM}$.

Different from the initial neuron, other common neurons aren't necessarily calculated from a single input. We take the calculation of neuron i's output as an example.

Assume that we have an existing random graph $\mathcal{G}$ and its two inner nodes $\nu_i, \nu_k$ satisfies $\nu_k\in\mathscr{A}\left(\nu_i\right)$. We will show the detailed calculation process of the corresponding neuron of $\nu_i$.

The sparse and compact features from neuron $k$ is denoted as $\textbf{F}_k = \left[\textbf{f}_{1,k},\cdots,\textbf{f}_{N_s,k}\right]\subseteq\mathbb{R}^{N\times N_s}$. Accordingly, the $d$-dimension approximation for initial neuron can be formulated as follows.
\begin{align}\label{eq:CN4}
\begin{split}
&\mathscr{F}_l\left(\textbf{f}_{z,k}\right) = \frac{1}{\sqrt{d}}\left(\sqrt{2}\cos\left(\omega^{\mathrm{T}}_{1,l}\textbf{f}_{z,k} + b_{1,l}\right),\cdots,\sqrt{2}\cos\left(\omega^{\mathrm{T}}_{d,l}\textbf{f}_{z,k} + b_{d,l}\right)\right)
\end{split}
\end{align}
where the weights $\omega_{m,l}$, $b_{m,l}, i = 1,2,\cdots,d$ are sampled from a given distributions $\mathscr{P}\left(\omega\right)$ and $U\left[0,2\pi\right]$, respectively.

Then we can obtain a FRF of neuron $i$ as follows.
\begin{align}\label{eq:CN5}
\begin{split}
\mathcal{F}^k_l = \left(\begin{array}{c}
\mathscr{F}_l\left(\textbf{f}_{1,k}\right)\\
\cdots\\
\mathscr{F}_l\left(\textbf{f}_{N,k}\right)\\
\end{array}\right)\subseteq\mathbb{R}^{N\times d}
\end{split}
\end{align}

The composite FRFs from neuron $k$ is formulated as $\mathcal{F}^k = \left[\mathcal{F}^k_1,\cdots,\mathcal{F}^k_M\right]$. And we can obtain the final output of the neuron $i$ $\mathscr{N}_i = \sum\limits_{k\in\mathcal{N}_i}\varphi\left(\mathcal{F}^kW_{n_i} + \beta_{n_i}\right)$. Then the output of the neurons wired under the guidance of graph $j$ is $\left[\left\{\mathscr{N}^j_{\mathcal{Q}}\right\}|\left\{\varphi\left(\mathscr{N}^j_{\mathcal{Q}}\bar{W}^j_{\mathcal{Q}} + \bar{\beta}^j_{\mathcal{Q}}\right)\right\}\right]$, where $\mathcal{Q}$ denotes a random permutations of natural numbers from $1$ to $\mathscr{X}_j$.

\subsection{EMA-ADMM Optimizer}

Next, we will introduce the weight optimization algorithm matched by RGNN. The core idea of this optimization algorithm is inherited from the traditional ADMM optimizer. The training process of a learning model is to find the optimal connection weights that make the network prediction output as close as possible to the training label. Meanwhile, it is also important to avoid overfitting in the training process. In that cases, pure mean square error (MSE) functions are not available at this point to help. This dilemma prompts the launch of the regularization item in the cost function to some extent. Aiming at solving this kind of convex programming, we usually introduce auxiliary variables to facilitate it. 

The traditional ADMM method is designed for general decomposition methods and decentralized algorithms in convex programming with equality constraints in reality. Varying from the previous ADMM optimizer, we take advantage of the Exponential Moving Average (EMA) techniques to improve the test accuracy, with enhanced robustness of the model. It can effectively avoid generating weight outliers in a single iteration.

Define the output of graph $i$ as $\mathcal{R} _i,i=1,2,\cdots ,\mathfrak{m}$. The connection weight can be written as $\left[ \boldsymbol{w}_{1}^{\mathrm{T}},\boldsymbol{w}_{2}^{\mathrm{T}},\cdots ,\boldsymbol{w}_{\mathfrak{m}}^{\mathrm{T}} \right] ^{\mathrm{T}}$. The output of an entire RGNN model is defined as follows.
\begin{align}
\begin{split}
\boldsymbol{Y}&=\mathcal{R} _1\boldsymbol{w}_1+\mathcal{R} _2\boldsymbol{w}_2+\cdots +\mathcal{R} _{\mathfrak{m}}\boldsymbol{w}_{\mathfrak{m}}\\
&=\boldsymbol{A}\left[ \boldsymbol{w}_{1}^{\mathrm{T}},\boldsymbol{w}_{2}^{\mathrm{T}},\cdots ,\boldsymbol{w}_{\mathfrak{m}}^{\mathrm{T}} \right] ^{\mathrm{T}}\\
\end{split}
\end{align}
where $\boldsymbol{A}$ is a pattern matrix similar to the fully connection matrix in deep learning. It comprise all the features we have extracted through a series of operations mentioned above. 

Let us define a new variable $\boldsymbol{W}$ representing the matrix $\left[ \boldsymbol{w}_{1}^{\mathrm{T}},\boldsymbol{w}_{2}^{\mathrm{T}},\cdots ,\boldsymbol{w}_{\mathfrak{m}}^{\mathrm{T}} \right] ^{\mathrm{T}}$. We can propose a primal problem that is similar to the optimization problem of SAE:
\begin{align}
\begin{split}
\mathrm{arg}\min_{\boldsymbol{W}} ||\boldsymbol{Y}-\boldsymbol{AW}||_{2}^{2}+||\boldsymbol{W}||_{2}^{2}
\end{split}
\end{align}

Different to the ADMM algorithms in previous works, we extra employ the EMA techniques to realize a fast convergence of optimization in network training. Although we have shown its brief description and calculation principle in the previous part, the specific iterative solving steps in each RGNN training will be exposed in the following.

According to the transformation of optimization problem defined in (13), the EMA-ADMM optimizer is designed in (14). Accordingly, different to the conventional ADMM optimizer, the result of any step in (14) will affect the following calculation process. In the last several steps, the connection weight of RGNN will chatter at the best point, thus we take the average value of it to make the NM more robust.

\begin{align}
\begin{split}
\begin{array}{c}
\left\{ \begin{array}{l}
W_{k+1}:=\left( Z^{\mathrm{T}}Z+\rho I \right) ^{-1}\left( Z^{\mathrm{T}}X+\rho \left( \mathcal{O} _k-u_k \right) \right)\\
W_{k+1}:=\left( 2W_{k+1}+kW_k \right) /\left( k+2 \right) ,\mathrm{if}\enspace k\ge 2\\
\end{array} \right.\\
\left\{ \begin{array}{l}
\mathcal{O} _{k+1}:=S_{\frac{\lambda}{\rho}}\left( W_{k+1}+u_k \right)\\
\mathcal{O} _{k+1}:=\left( 2\mathcal{O} _{k+1}+k\mathcal{O} _k \right) /\left( k+2 \right) ,\mathrm{if}\enspace k\ge 2\\
\end{array} \right.\\
\left\{ \begin{array}{l}
u_{k+1}:=u_k+\left( W_{k+1}-\mathcal{O} _{k+1} \right)\\
u_{k+1}:=\left( 2u_{k+1}+ku_k \right) /\left( k+2 \right) ,\mathrm{if}\enspace k\ge 2\\
\end{array} \right.\\
\end{array}
\end{split}
\end{align}

\subsection{Stochastic Neuromorphic Completeness}

According to the definition of neuromorphic completeness in \upcite{zhang2020system}, for arbitrary given error $\epsilon$ and any Turing-computable function $f\left(x\right)$, a computational system is called neuromorphic complete if it can establish a new function $F\left(x\right)$  which satisfies $||F\left(x\right) - f\left(x\right)|| < \epsilon$. This definition allows an algorithm to approximate a function instead of exactly computing it that is similar to the universal approximation theorem for general NMs. Different from the general NMs designed by users, the RGNN model uses random graphs to design its inner architecture. And how to guarantee its availability in dealing with data mining under arbitrary $p$ is also a proposition to be solved in urgent need. Motivated by the definition of neuromorphic completeness, we define a concept termed stochastic neuromorphic completeness. It is a vital property of RGNN and a new distance variable $\rho$ is defined to prove it. Let us define some mathematical symbols first. Consider any continuous function $f\left( x \right) \in C\left( \boldsymbol{I}^d \right)$ , defined on the compact set of the standard hypercube $\boldsymbol{K}\subset \boldsymbol{I}^d=\left[ 0;1 \right]$. The distance $\rho$ between the approximation function $f_a\left(x\right)$ and function $f\left( x \right)$ on the compact set $\boldsymbol{K}$ can be defined as:
\begin{equation}
\rho _{\boldsymbol{K}}\left( f,f_a \right) =\sqrt{\mathbb{E} \left[ \int_{\boldsymbol{K}}{\left( f\left( x \right) -f_a\left( x \right) \right) ^2}dx \right]}
\end{equation}

The definition of stochastic neuromorphic completeness is expressed as follow.

\textbf{Definition 1}: For any arbitrary given error $\epsilon$, connection probability $p$ and any continuous function $f\left( x \right) \in C\left( \boldsymbol{I}^d \right)$, if we find a computational system which can calculate an approximation function $f_a\left( x \right)$ satisfying $\lim_{a\rightarrow \infty} \rho _{\boldsymbol{K}}\left( f,f_a \right) <\epsilon $, we can say that this computational system has a property of stochastic neuromorphic completeness.
Thus we have the following results.

\textbf{Theorem 1}:For any compact set and arbitrary functions $f\left( x \right) \in C\left( \boldsymbol{I}^d \right)$, the RGNN model can compute corresponding approximation function $f_a\left( x \right)$ which satisfies $\lim_{a\rightarrow \infty} \rho _{\boldsymbol{K}}\left( f,f_a \right) =0$.

This result has parallels in the universal approximation capability proof of SNNs, yet universal approximation capability is more suitable for NN models with fixed structure. In view of the structural uncertainty proposed in our research, we can not derive a proven availability for RGNN in data mining using universal approximation theorem. According to the Theorem 1, we can reach the following ratiocination: for arbitrary given error $\epsilon$ and any continuous function $f\left( x \right) \in C\left( \boldsymbol{I}^d \right)$, the random computational system RGNN with any connection probability $p\in\left(0,1\right)$ can calculate an approximation function $f_a\left(x\right)$ that satisfies $\lim_{a\rightarrow \infty} \rho _{\boldsymbol{K}}\left( f,f_a \right) <\epsilon $. 

In essence, we propose stochastic neuromorphic completeness to ensure the capability of a single RGNN in a big data environment. It lays the theoretical foundation for the feasibility of randomly wired neural learning algorithms in image classification, linear regression, and data clustering.

\subsection*{Proof of Stochastic Neuromorphic Completeness}
In this section, we will illustrate a method to prove the stochastic neuromorphic completeness property of RGNN which has been stated in Theorem 1. Assume that the RGNN consists of $n$ graphs which contain $\mathscr{X} _1,\cdots ,\mathscr{X} _n$ neurons, respectively. And the $i$th graph can be denoted by three components as $\mathcal{G} _i\equiv \left( \mathcal{V} _i,\mathcal{E} _i,\mathcal{A} _i \right)$.The output of this graph is corresponding to $\mathscr{N} _{1}^{i},\cdots ,\mathscr{N} _{\mathscr{X} _i}^{i}$. Let us define $w_{initial}^{k},k=1,\cdots ,n$ denoting the weight matrix connecting the initial node of each graph to the output layer. Therefore, for any finite integer $n$, define:
\begin{align}\label{eq:PS1}
\begin{split}
&\boldsymbol{f}_{initial}=\sum_{k=1}^n{w_{initial}^{k}}\mathscr{N} _{1}^{k},\boldsymbol{f}_{initial}^{e}=\sum_{k=1}^n{\bar{w}_{initial}^{k}}\phi \left( \mathscr{N} _{1}^{k}\bar{W}_{1}^{k}+\bar{\beta}_{1}^{k} \right) 
\end{split}
\end{align}

It is obvious that the construction processes of the initial neurons in each graph only use bounded continuous function like cosine function, sigmoid function. And the inner weight matrix and bias are sampled from a given probability. We can infer that resident function defined as $\boldsymbol{f}_r=\boldsymbol{f}-\boldsymbol{f}_{initial}-\boldsymbol{f}_{initial}^{e}$ is also bounded and integrable in $\boldsymbol{I}^d$.Furthermore, there exists a function $\boldsymbol{f}_{c_n}\in C\left( \boldsymbol{I}^d \right) $ that satisfy:
\begin{align}\label{eq:PS2}
\begin{split}
\rho _K\left( \boldsymbol{f}_r,\boldsymbol{f}_{c_n} \right) <\frac{\varepsilon}{2},\forall \varepsilon >0
\end{split}
\end{align}

The conclusion in (\ref{eq:PS2}) could be theoretically guaranteed by the research findings in \upcite{igelnik1995stochastic}. In addition, for any $\boldsymbol{f}_{c_n}\in L^2\left( K \right) $, since other neurons are also established through nonconstant and bounded functions, we can use other neurons in these graphs to obtain a smooth function $\boldsymbol{f}_t$ such that $||\boldsymbol{f}_t-\boldsymbol{f}_{c_n}||<\epsilon ,\forall \epsilon >0$. For example, the function $\boldsymbol{f}_{t}^{i}$ is constructed by the neurons in graph $i$ except for initial neuron that has the following form:
\begin{align}\label{eq:PS3}
\begin{split}
\boldsymbol{f}_{t}^{i}=\sum_{j=2}^{\mathscr{X} _i}{w_j}\sum_{k\in \mathcal{N} _j}{\phi}\left( \mathcal{F} ^kW_{n_j}+\beta _{n_j} \right) 
&+\sum_{j=2}^{\mathscr{X} _i}{\bar{w}_j}\phi \left( \left( \sum_{k\in \mathcal{N} _j}{\phi}\left( \mathcal{F} ^kW_{n_j}+\beta _{n_j} \right) \right) \bar{W}_{j}^{i}+\bar{\beta}_{j}^{i} \right) 
\\
&\boldsymbol{f}_t=\sum_{c=1}^i{\boldsymbol{f}_{t}^{i}}
\end{split}
\end{align}
where $w_j,\bar{w}_j$ stand for the weight matrix connecting the neurons to the output layer.

In (\ref{eq:PS3}), the weight matrix and bias are all sampled from a specific probability measures. Thus according to the universal approximation property of RVFL (details please refer to\upcite{igelnik1995stochastic,pao1992functional}), we can derive 
\begin{align}\label{eq:PS4}
\begin{split}
\rho _K\left( \boldsymbol{f}_{c_n},\boldsymbol{f}_t \right) <\frac{\varepsilon}{2}
\end{split}
\end{align}

Performing some manipulation on the distance $\rho$ gives us:
\begin{align}\label{eq:PS5}
\begin{split}
&\rho _{\boldsymbol{K}}\left( \boldsymbol{f},\boldsymbol{f}_a \right) =\sqrt{\mathbb{E} \left[ \int_{\boldsymbol{K}}{\left( \boldsymbol{f}\left( x \right) -\boldsymbol{f}_a\left( x \right) \right) ^2}dx \right]}\\
&=\sqrt{\mathbb{E} \left[ \int_{\boldsymbol{K}}{\left( \boldsymbol{f}\left( x \right) -\boldsymbol{f}_{initial}-\boldsymbol{f}_{initial}^{e}-\boldsymbol{f}_t \right) ^2}dx \right]}\\
&=\sqrt{\mathbb{E} \left[ \int_{\boldsymbol{K}}{\left( \boldsymbol{f}_r-\boldsymbol{f}_t \right) ^2}dx \right]}\le \sqrt{\mathbb{E} \left[ \int_{\boldsymbol{K}}{\left( \boldsymbol{f}_r-\boldsymbol{f}_{c_n} \right) ^2}dx \right]}+\sqrt{\mathbb{E} \left[ \int_{\boldsymbol{K}}{\left( \boldsymbol{f}_{c_n}-\boldsymbol{f}_t \right) ^2}dx \right]}\\
&=\varepsilon\\
\end{split}
\end{align}

According to the theorem and analysis in\upcite{igelnik1995stochastic,pao1992functional,mao2021broad,chen2018universal}, we can derive
\begin{align}\label{eq:PS6}
\begin{split}
\lim_{\mathscr{X} _1,\cdots ,\mathscr{X} _n\rightarrow \infty} \rho _{\boldsymbol{K}}\left( \boldsymbol{f},\boldsymbol{f}_a \right) =0
\end{split}
\end{align}

Thus we complete the proof of theorem 1 that states the stochastic neuromorphic completeness property of RGNN.

\subsection{Joint Classification in M-RGNNs}

To fully tap the diversity of NNs generated through random graphs and aiming at completing classification task,  how to use the prediction results of each RGNN in M-RGNNs is vital for autonomous classification without manual intervention. Here we show the details of the process dealing with prediction results from different RGNNs.
\begin{enumerate}
	\item Sample a probability variable $p$ and use this probability variable to generate a random graph.
	\item Use the graph generated in the last step to establish a RGNN and employ the EMA-ADMM to train it.
	\item Determine whether the number of RGNNs reaches the setting value. If it doesn't reach the setting value, jump to step 1, otherwise continue to step 4.
	\item All the RGNNs in M-RGNNs are used to classify the data and the prediction results of each RGNN are recorded.
	\item The highest frequency category in the prediction results is counted and identified as the final prediction result of in M-RGNNs.
\end{enumerate}

This mechanism is equivalent to empowering the framework to determine its own structure autonomously. Additionally, the final prediction results can be obtained based on the principle of minority obeying majority through comparing the prediction results of all the self-generated RGNNs.

\section*{Acknowledgements}
The authors are very thankful to the Yann LeCun, Dr. William H. Wolberg, the National Institute of Diabetes Digestive and Kidney Diseases, AT\&T Laboratories Cambridge, Kuang-Chih Lee, Graham, Chinese University of HongKong, Zalando Research, James A. Haley Veterans’ Hospital, University of South Florida and Moffitt Cancer Center for opening the access to the MNIST dataset, NORB dataset, WBC dataset, PID dataset, Extended Yaleb dataset, UMIST dataset, ORL dataset, CelebA dataset, Fashion MNIST dataset, lung and colon cancer histopathological images and allowing them to be used.

\noindent\textbf{Funding:} National Natural Science Foundation of China (NSFC) under Grant U1813225;\\
National Key Research and Development Project under Grant 2019YFB1310404.

\section*{Author Contribution}
\noindent Conceptualization: Rongxin Cui, Ruiqi Mao\\
Methodology: Ruiqi Mao, Rongxin Cui\\
Investigation: Rongxin Cui\\
Visualization: Ruiqi Mao, Rongxin Cui\\
Funding acquisition: Rongxin Cui\\
Project administration: Rongxin Cui\\
Supervision: Rongxin Cui\\
Writing – original draft: Ruiqi Mao\\
Writing – review \& editing: Ruiqi Mao, Rongxin Cui\\

\section*{Competing Interests}
Authors declare that they have no competing interests.
\section*{Data and materials availability}
All data are available in the main text or the supplementary materials.

\bibliographystyle{Bibliography/IEEEtranTIE}
\bibliography{Bibliography/IEEEabrv,Bibliography/BIB_xx-TIE-xxxx}\ 

\begin{thebibliography}{10}
\providecommand{\url}[1]{#1}
\csname url@samestyle\endcsname
\providecommand{\newblock}{\relax}
\providecommand{\bibinfo}[2]{#2}
\providecommand{\BIBentrySTDinterwordspacing}{\spaceskip=0pt\relax}
\providecommand{\BIBentryALTinterwordstretchfactor}{4}
\providecommand{\BIBentryALTinterwordspacing}{\spaceskip=\fontdimen2\font plus
\BIBentryALTinterwordstretchfactor\fontdimen3\font minus
  \fontdimen4\font\relax}
\providecommand{\BIBforeignlanguage}[2]{{%
\expandafter\ifx\csname l@#1\endcsname\relax
\typeout{** WARNING: IEEEtran.bst: No hyphenation pattern has been}%
\typeout{** loaded for the language `#1'. Using the pattern for}%
\typeout{** the default language instead.}%
\else
\language=\csname l@#1\endcsname
\fi
#2}}
\providecommand{\BIBdecl}{\relax}
\BIBdecl

\bibitem{2020Magnetic}
H.~Y. Kwon, H.~G. Yoon, C.~Lee, G.~Chen, and C.~Won, ``Magnetic hamiltonian
  parameter estimation using deep learning techniques,'' \emph{Science
  Advances}, vol.~6, no.~39, p. eabb0872.

\bibitem{ham2019deep}
Y.~G. Ham, J.~H. Kim, and J.~J. Luo, ``Deep learning for multi-year enso
  forecasts,'' \emph{Nature}, vol. 573, no. 7775, pp. 568--572, 2019.

\bibitem{chiu2021predicting}
Y.~C. Chiu, S.~Zheng, L.~Wang, B.~S. Iskra, M.~K. Rao, P.~J. Houghton,
  Y.~Huang, and Y.~Chen, ``Predicting and characterizing a cancer dependency
  map of tumors with deep learning,'' \emph{Science Advances}, vol.~7, no.~34,
  p. eabh1275, 2021.

\bibitem{baek2021accurate}
M.~Baek, F.~DiMaio, I.~Anishchenko, J.~Dauparas, S.~Ovchinnikov, G.~R. Lee,
  J.~Wang, Q.~Cong, L.~N. Kinch, R.~D. Schaeffer \emph{et~al.}, ``Accurate
  prediction of protein structures and interactions using a three-track neural
  network,'' \emph{Science}, vol. 373, no. 6557, pp. 871--876, 2021.

\bibitem{cheng2021robust}
S.~Cheng, S.~Liu, J.~Yu, G.~Rao, Y.~Xiao, W.~Han, W.~Zhu, X.~Lv, N.~Li, J.~Cai
  \emph{et~al.}, ``Robust whole slide image analysis for cervical cancer
  screening using deep learning,'' \emph{Nature communications}, vol.~12,
  no.~1, pp. 1--10, 2021.

\bibitem{warnat2021swarm}
S.~Warnat-Herresthal, H.~Schultze, K.~L. Shastry, S.~Manamohan, S.~Mukherjee,
  V.~Garg, R.~Sarveswara, K.~H{\"a}ndler, P.~Pickkers, N.~A. Aziz
  \emph{et~al.}, ``Swarm learning for decentralized and confidential clinical
  machine learning,'' \emph{Nature}, vol. 594, no. 7862, pp. 265--270, 2021.

\bibitem{silver2017mastering}
D.~Silver, J.~Schrittwieser, K.~Simonyan, I.~Antonoglou, A.~Huang, A.~Guez,
  T.~Hubert, L.~Baker, M.~Lai, A.~Bolton \emph{et~al.}, ``Mastering the game of
  go without human knowledge,'' \emph{nature}, vol. 550, no. 7676, pp.
  354--359, 2017.

\bibitem{havlivcek2019supervised}
V.~Havl{\'\i}{\v{c}}ek, A.~D. C{\'o}rcoles, K.~Temme, A.~W. Harrow, A.~Kandala,
  J.~M. Chow, and J.~M. Gambetta, ``Supervised learning with quantum-enhanced
  feature spaces,'' \emph{Nature}, vol. 567, no. 7747, pp. 209--212, 2019.

\bibitem{kavran2021denoising}
A.~J. Kavran and A.~Clauset, ``Denoising large-scale biological data using
  network filters,'' \emph{BMC bioinformatics}, vol.~22, no.~1, pp. 1--21,
  2021.

\bibitem{jia2021unifying}
J.~Jia and A.~R. Benson, ``A unifying generative model for graph learning
  algorithms: Label propagation, graph convolutions, and combinations,''
  \emph{arXiv preprint arXiv:2101.07730}, 2021.

\bibitem{jia2020residual}
J.~Jia and A.~R. Benson, ``Residual correlation in graph neural network
  regression,'' in \emph{Proceedings of the 26th ACM SIGKDD International
  Conference on Knowledge Discovery \& Data Mining}, pp. 588--598, 2020.

\bibitem{mignan2019one}
A.~Mignan and M.~Broccardo, ``One neuron versus deep learning in aftershock
  prediction,'' \emph{Nature}, vol. 574, no. 7776, pp. E1--E3, 2019.

\bibitem{zhang2018predicting}
J.~Zhang, Y.~Zheng, D.~Qi, R.~Li, X.~Yi, and T.~Li, ``Predicting citywide crowd
  flows using deep spatio-temporal residual networks,'' \emph{Artificial
  Intelligence}, vol. 259, pp. 147--166, 2018.

\bibitem{he2016deep}
K.~He, X.~Zhang, S.~Ren, and J.~Sun, ``Deep residual learning for image
  recognition,'' in \emph{Proceedings of the IEEE conference on computer vision
  and pattern recognition}, pp. 770--778, 2016.

\bibitem{gang2021recognition}
L.~Gang, Z.~Haixuan, E.~Linning, Z.~Ling, L.~Yu, and Z.~Juming, ``Recognition
  of honeycomb lung in ct images based on improved mobilenet model,''
  \emph{Medical Physics}, vol.~48, no.~8, pp. 4304--4315, 2021.

\bibitem{sandler2018mobilenetv2}
M.~Sandler, A.~Howard, M.~Zhu, A.~Zhmoginov, and L.~C. Chen, ``Mobilenetv2:
  Inverted residuals and linear bottlenecks,'' in \emph{Proceedings of the IEEE
  conference on computer vision and pattern recognition}, pp. 4510--4520, 2018.

\bibitem{xie2019exploring}
S.~Xie, A.~Kirillov, R.~Girshick, and K.~He, ``Exploring randomly wired neural
  networks for image recognition,'' in \emph{Proceedings of the IEEE/CVF
  International Conference on Computer Vision}, pp. 1284--1293, 2019.

\bibitem{mou2021analog}
X.~Mou, J.~Tang, Y.~Lyu, Q.~Zhang, S.~Yang, F.~Xu, W.~Liu, M.~Xu, Y.~Zhou,
  W.~Sun \emph{et~al.}, ``Analog memristive synapse based on topotactic phase
  transition for high-performance neuromorphic computing and neural network
  pruning,'' \emph{Science Advances}, vol.~7, no.~29, p. eabh0648, 2021.

\bibitem{gutig2016spiking}
R.~G{\"u}tig, ``Spiking neurons can discover predictive features by
  aggregate-label learning,'' \emph{Science}, vol. 351, no. 6277, p. aab4113,
  2016.

\bibitem{zhang2021self}
T.~Zhang, X.~Cheng, S.~Jia, M.~m. Poo, Y.~Zeng, and B.~Xu,
  ``Self-backpropagation of synaptic modifications elevates the efficiency of
  spiking and artificial neural networks,'' \emph{Science Advances}, vol.~7,
  no.~43, p. eabh0146, 2021.

\bibitem{noe2019boltzmann}
F.~No{\'e}, S.~Olsson, J.~K{\"o}hler, and H.~Wu, ``Boltzmann generators:
  Sampling equilibrium states of many-body systems with deep learning,''
  \emph{Science}, vol. 365, no. 6457, p. eaaw1147, 2019.

\bibitem{kaufmann2020crystal}
K.~Kaufmann, C.~Zhu, A.~S. Rosengarten, D.~Maryanovsky, T.~J. Harrington,
  E.~Marin, and K.~S. Vecchio, ``Crystal symmetry determination in electron
  diffraction using machine learning,'' \emph{Science}, vol. 367, no. 6477, pp.
  564--568, 2020.

\bibitem{luo2021ecnet}
Y.~Luo, G.~Jiang, T.~Yu, Y.~Liu, L.~Vo, H.~Ding, Y.~Su, W.~W. Qian, H.~Zhao,
  and J.~Peng, ``Ecnet is an evolutionary context-integrated deep learning
  framework for protein engineering,'' \emph{Nature communications}, vol.~12,
  no.~1, pp. 1--14, 2021.

\bibitem{hsieh2021automated}
C.~I. Hsieh, K.~Zheng, C.~Lin, L.~Mei, L.~Lu, W.~Li, F.~P. Chen, Y.~Wang,
  X.~Zhou, F.~Wang \emph{et~al.}, ``Automated bone mineral density prediction
  and fracture risk assessment using plain radiographs via deep learning,''
  \emph{Nature communications}, vol.~12, no.~1, pp. 1--9, 2021.

\bibitem{lin2018all}
X.~Lin, Y.~Rivenson, N.~T. Yardimci, M.~Veli, Y.~Luo, M.~Jarrahi, and A.~Ozcan,
  ``All-optical machine learning using diffractive deep neural networks,''
  \emph{Science}, vol. 361, no. 6406, pp. 1004--1008, 2018.

\bibitem{robert1989theory}
H.-N. Robert \emph{et~al.}, ``Theory of the backpropagation neural network,''
  \emph{Proc. 1989 IEEE IJCNN}, vol.~1, pp. 593--605, 1989.

\bibitem{shi2014linear}
W.~Shi, Q.~Ling, K.~Yuan, G.~Wu, and W.~Yin, ``On the linear convergence of the
  admm in decentralized consensus optimization,'' \emph{IEEE Transactions on
  Signal Processing}, vol.~62, no.~7, pp. 1750--1761, 2014.

\bibitem{shen2012distributed}
C.~Shen, T.~H. Chang, K.~Y. Wang, Z.~Qiu, and C.~Y. Chi, ``Distributed robust
  multicell coordinated beamforming with imperfect csi: An admm approach,''
  \emph{IEEE Transactions on signal processing}, vol.~60, no.~6, pp.
  2988--3003, 2012.

\bibitem{wang2019global}
Y.~Wang, W.~Yin, and J.~Zeng, ``Global convergence of admm in nonconvex
  nonsmooth optimization,'' \emph{Journal of Scientific Computing}, vol.~78,
  no.~1, pp. 29--63, 2019.

\bibitem{lecun2004learning}
Y.~LeCun, F.~J. Huang, and L.~Bottou, ``Learning methods for generic object
  recognition with invariance to pose and lighting,'' in \emph{Proceedings of
  the 2004 IEEE Computer Society Conference on Computer Vision and Pattern
  Recognition, 2004. CVPR 2004.}, vol.~2, pp. II--104.\hskip 1em plus 0.5em
  minus 0.4em\relax IEEE, 2004.

\bibitem{lecun1998gradient}
Y.~LeCun, L.~Bottou, Y.~Bengio, and P.~Haffner, ``Gradient-based learning
  applied to document recognition,'' \emph{Proc. IEEE}, vol.~86, no.~11, pp.
  2278--2324, 1998.

\bibitem{smith1988using}
J.~W. Smith, J.~E. Everhart, W.~Dickson, W.~C. Knowler, and R.~S. Johannes,
  ``Using the adap learning algorithm to forecast the onset of diabetes
  mellitus,'' in \emph{Proceedings of the annual symposium on computer
  application in medical care}, pp. 261--265.\hskip 1em plus 0.5em minus
  0.4em\relax American Medical Informatics Association, 1988.

\bibitem{WBC}
\BIBentryALTinterwordspacing
W.~H. Wolberg, W.~N. Street, and O.~L. Mangasarian, ``Breast cancer wisconsin
  data set.'' [Online]. Available:
  \url{https://archive.ics.uci.edu/ml/datasets/Breast+Cancer+Wisconsin+\%28Diagnostic\%29}
\BIBentrySTDinterwordspacing

\bibitem{xiao2017fashion}
H.~Xiao, K.~Rasul, and R.~Vollgraf, ``Fashion-mnist: a novel image dataset for
  benchmarking machine learning algorithms,'' \emph{arXiv preprint
  arXiv:1708.07747}, 2017.

\bibitem{umist}
\BIBentryALTinterwordspacing
D.~Graham, ``Umist face database.'' [Online]. Available:
  \url{http://images.ee.umist.ac.uk/danny/database.html}
\BIBentrySTDinterwordspacing

\bibitem{lee2005acquiring}
K.~C. Lee, J.~Ho, and D.~J. Kriegman, ``Acquiring linear subspaces for face
  recognition under variable lighting,'' \emph{IEEE Transactions on pattern
  analysis and machine intelligence}, vol.~27, no.~5, pp. 684--698, 2005.

\bibitem{ORL}
\BIBentryALTinterwordspacing
A.~L. Cambridge, ``The orl database.'' [Online]. Available:
  \url{http://www.cl.cam.ac.uk/Research/DTG/attarchive}
\BIBentrySTDinterwordspacing

\bibitem{liu2015deep}
Z.~Liu, P.~Luo, X.~Wang, and X.~Tang, ``Deep learning face attributes in the
  wild,'' in \emph{Proceedings of the IEEE international conference on computer
  vision}, pp. 3730--3738, 2015.

\bibitem{Cancer}
\BIBentryALTinterwordspacing
A.~A. Borkowski, M.~M. Bui, L.~B. Thomas, C.~P. Wilson, L.~A. DeLand, and S.~M.
  Mastorides, ``Lc25000 lung and colon histopathological image dataset.''
  [Online]. Available: \url{https://github.com/tampapath/lung_colon_image_set}
\BIBentrySTDinterwordspacing

\bibitem{zhang2020system}
Y.~Zhang, P.~Qu, Y.~Ji, W.~Zhang, G.~Gao, G.~Wang, S.~Song, G.~Li, W.~Chen,
  W.~Zheng \emph{et~al.}, ``A system hierarchy for brain-inspired computing,''
  \emph{Nature}, vol. 586, no. 7829, pp. 378--384, 2020.

\bibitem{igelnik1995stochastic}
B.~Igelnik and Y.-H. Pao, ``Stochastic choice of basis functions in adaptive
  function approximation and the functional-link net,'' \emph{IEEE transactions
  on Neural Networks}, vol.~6, no.~6, pp. 1320--1329, 1995.

\bibitem{pao1992functional}
Y.-H. Pao and Y.~Takefuji, ``Functional-link net computing: theory, system
  architecture, and functionalities,'' \emph{IEEE Computer}, vol.~25, no.~5,
  pp. 76--79, 1992.

\bibitem{mao2021broad}
R.~Mao, R.~Cui, and C.~P. Chen, ``Broad learning with reinforcement learning
  signal feedback: Theory and applications,'' \emph{IEEE Transactions on Neural
  Networks and Learning Systems}, 2021.

\bibitem{chen2018universal}
C.~P. Chen, Z.~Liu, and S.~Feng, ``Universal approximation capability of broad
  learning system and its structural variations,'' \emph{IEEE transactions on
  neural networks and learning systems}, vol.~30, no.~4, pp. 1191--1204, 2018.

\end{thebibliography}



\end{document}